\documentclass[conference]{IEEEtran}
\usepackage{amsmath,amsfonts}
\usepackage{algorithmic}
\usepackage{algorithm}
\usepackage{array}
\usepackage[caption=false,font=normalsize,labelfont=sf,textfont=sf]{subfig}
\usepackage{textcomp}
\usepackage{stfloats}
\usepackage{url}
\usepackage{verbatim}
\usepackage{graphicx}
\usepackage{cite}
\usepackage{booktabs}
\usepackage[hidelinks]{hyperref}
\begin{document}

\title{Probabilistic Sensing: Intelligence in Data Sampling}

\author{Ibrahim Albulushi\textsuperscript{1$\dagger$}, Saleh Bunaiyan\textsuperscript{1,2$\dagger$}, Suraj S. Cheema\textsuperscript{3}, Hesham ElSawy\textsuperscript{4}, and Feras Al-Dirini\textsuperscript{3,4*}\\ \textsuperscript{1}EE, KFUPM, Dhahran, KSA, \textsuperscript{2}ECE, UCSB, Santa Barbara, CA, USA, \textsuperscript{3}RLE, MIT, Cambridge, MA, USA, \\  \textsuperscript{4}School of Computing, Queen's University, Kingston, ON, Canada, \\  $^\dagger$equally contributing authors, *email: {\underline{aldirini@mit.edu}}}

\markboth{Journal of XXXXXXXXX,~Vol.~XX, No.~XX, XXX-XXXX}%
{Shell \MakeLowercase{\textit{et al.}}: A Sample Article Using IEEEtran.cls for IEEE Journals}

\maketitle

\begin{abstract}
Extending the intelligence of sensors to the data-acquisition process - deciding whether to sample or not - can result in transformative energy-efficiency gains. However, making such a decision in a deterministic manner involves risk of losing information. Here we present a sensing paradigm that enables making such a decision in a probabilistic manner. The paradigm takes inspiration from the autonomous nervous system and employs a probabilistic neuron (p-neuron) driven by an analog feature extraction circuit. The response time of the system is on the order of microseconds, over-coming the sub-sampling-rate response time limit and enabling real-time intelligent autonomous activation of data-sampling. Validation experiments on active seismic survey data demonstrate lossless probabilistic data acquisition, with a normalized mean squared error of 0.41$\%$, and 93$\%$ saving in the active operation time of the system and the number of generated samples. 

\end{abstract}


\section{Introduction}
\IEEEPARstart{T}{he} increasing inflation in the amount of generated data by sensors and devices is surpassing the capacity of computing and artificial intelligence systems, making it very difficult to entirely process such data. This triggers a need for greater intelligence - not only in processing of sensory data - but also in the generation and acquisition of such data in the first place. 

Many approaches have attempted achieving intelligence in data acquisition through event-based sensing, 
where at the occurrence of an event of interest - identified through real-time analog event detection - specific features of the event signal \cite{wasted_power,Feature_4,IAF2,AFE,AFE2}, or even the event signal itself \cite{our_paper, Sensor_Journal}, are selectively sampled. Such an approach entails a critical decision that cannot be undone; whether to sample or not. While such a decision seems to be deterministic, it is indeed probabilistic. Here we present a sensing paradigm that employs a probabilistic neuron (p-neuron) \cite{PatentPNeurons}, implemented using a probabilistic bit (p-bit) \cite{PatentDecoupledPBit,PatentTunablePBit}, that enables making such a decision in a probabilistic manner\cite{PatentPSensing}; achieving real-time event-based probabilistic autonomous activation of data sampling. 

\begin{figure}[!t]
\includegraphics[width=0.95\columnwidth]{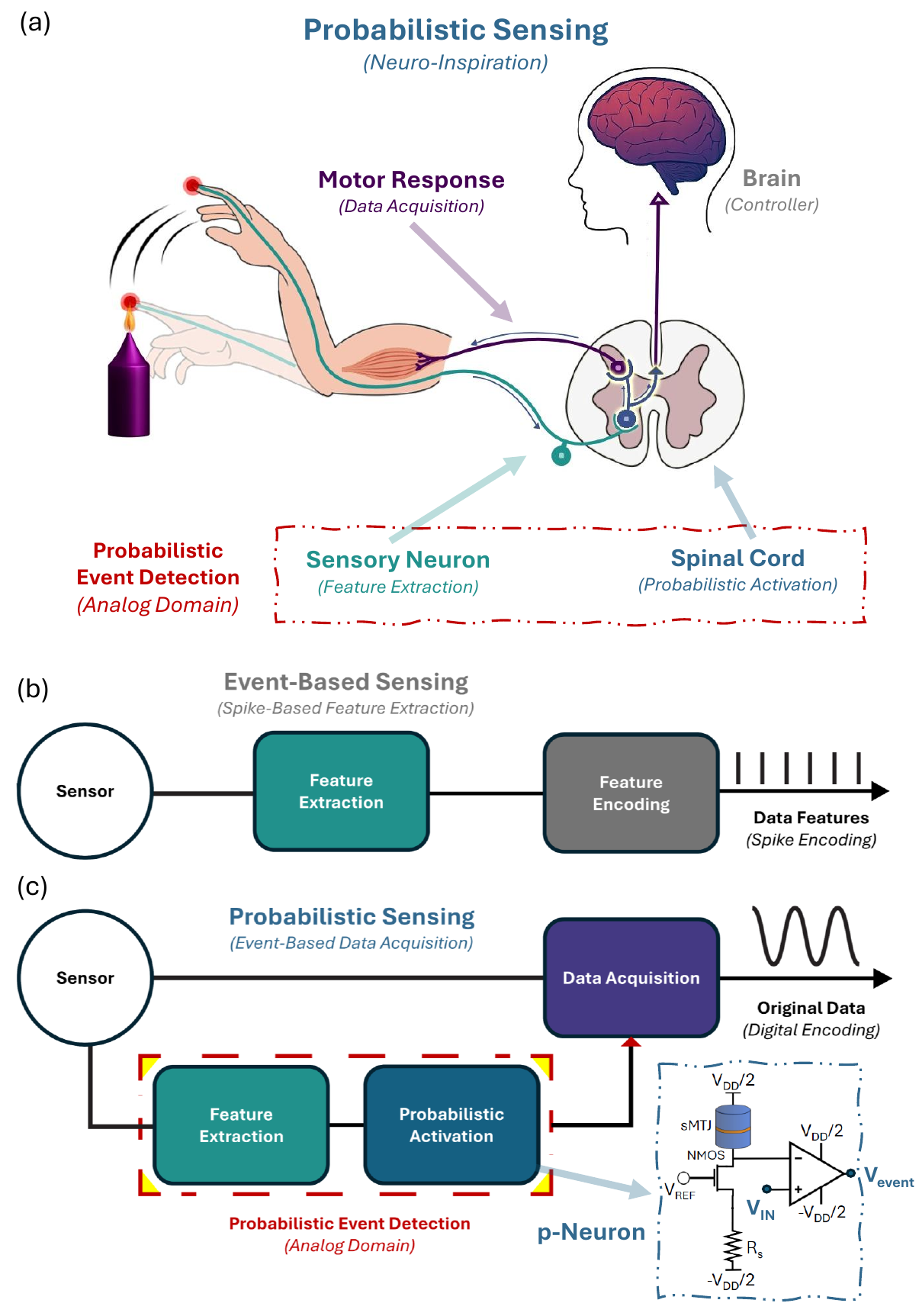}
    \centering
    \caption{\textbf{Probabilistic sensing using p-neurons.} (a) The neuro-inspiration behind probabilistic sensing. (b) Conventional event-based sensing, where event features are extracted from the sensor signal and then these features are encoded into spikes or spike trains. (c) Probabilistic sensing, where the extracted event features from the sensor signal are used to activate a probabilistic neuron (p-neuron), which in turn activates data-acquisition, capturing the actual sensor signal and retaining the original data, not just a few features of it. 
    The inset shows a spintronic p-neuron implemented using a probabilistic bit (p-bit).  
    }
    \label{Neuro}
\end{figure}

The neuro-inspiration behind this paradigm is illustrated in Fig. \ref{Neuro} (a). In the autonomous nervous system, sensory stimuli can activate an instantaneous autonomous motor response, known as a reflex, without consulting the brain. The sensory neuron is continuously monitoring sensory stimuli, while the brain is completely unaware. The sensory neuron function is achieved with a probabilistic event-detection unit (employing a p-neuron), while the triggered autonomous reflex response is the activation of all system blocks, including the data-acquisition unit, as illustrated in Fig. \ref{Neuro} (c).

\begin{figure*}[!t]
\includegraphics[width=6.6 in]{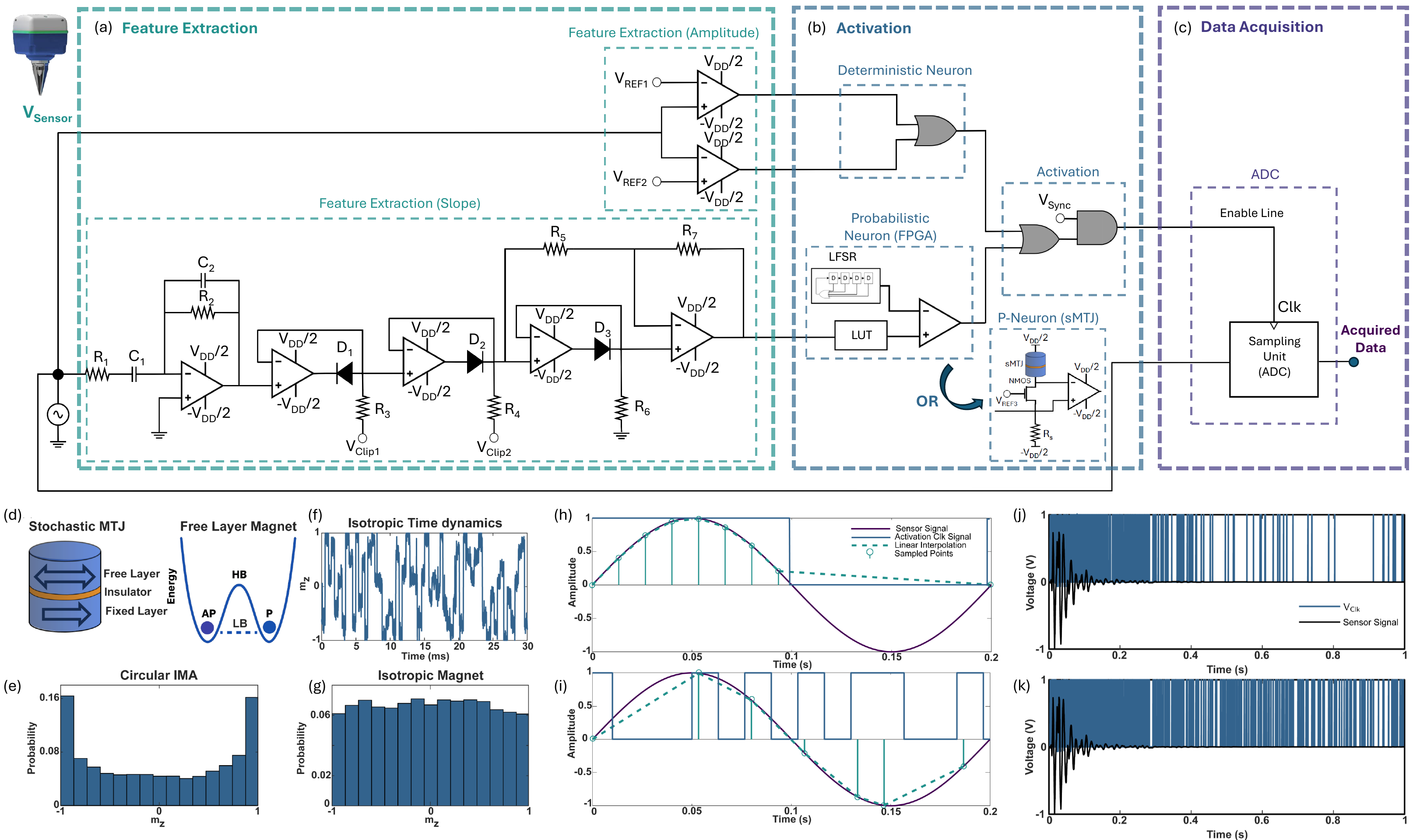}
    \centering
    \caption{\textbf{Overall probabilistic sensing system design}, including the (a) feature extraction, (b) activation, and (c) data acquisition units. The p-neuron is in the activation unit, shown here with two different implementations; digital (top) and spintronic (bottom). The entropy source in the digital implementation is a Linear Feedback Shift Register (LFSR) and in the spintronic implementation is a stochastic MTJ. \textbf{Magnetization dynamics of the sMTJ and its impact on probabilistic sampling.} (d) A stochastic MTJ (sMTJ) with a low-barrier (LB) free layer magnet. Magnetization dynamics of $m_z$ for (e) a circular in-plane magnetic anisotropy (IMA) magnet, and (f)-(g) an isotropic magnet. Sensor signal sampling with the sMTJ retention time (h) longer than the sampling time and (i) comparable to the sampling time. A seismic geophone senosor signal and the corresponding event-detection signal obtain from the activation unit ($V_{Clk}$), where in (j) the average random sampling rate (X) is smaller than its value in (k). 
    }
    \label{MTJ}
\end{figure*}

\section{Overall System Architecture}

\begin{figure*}[!t]
\includegraphics[width= 6.6 in]{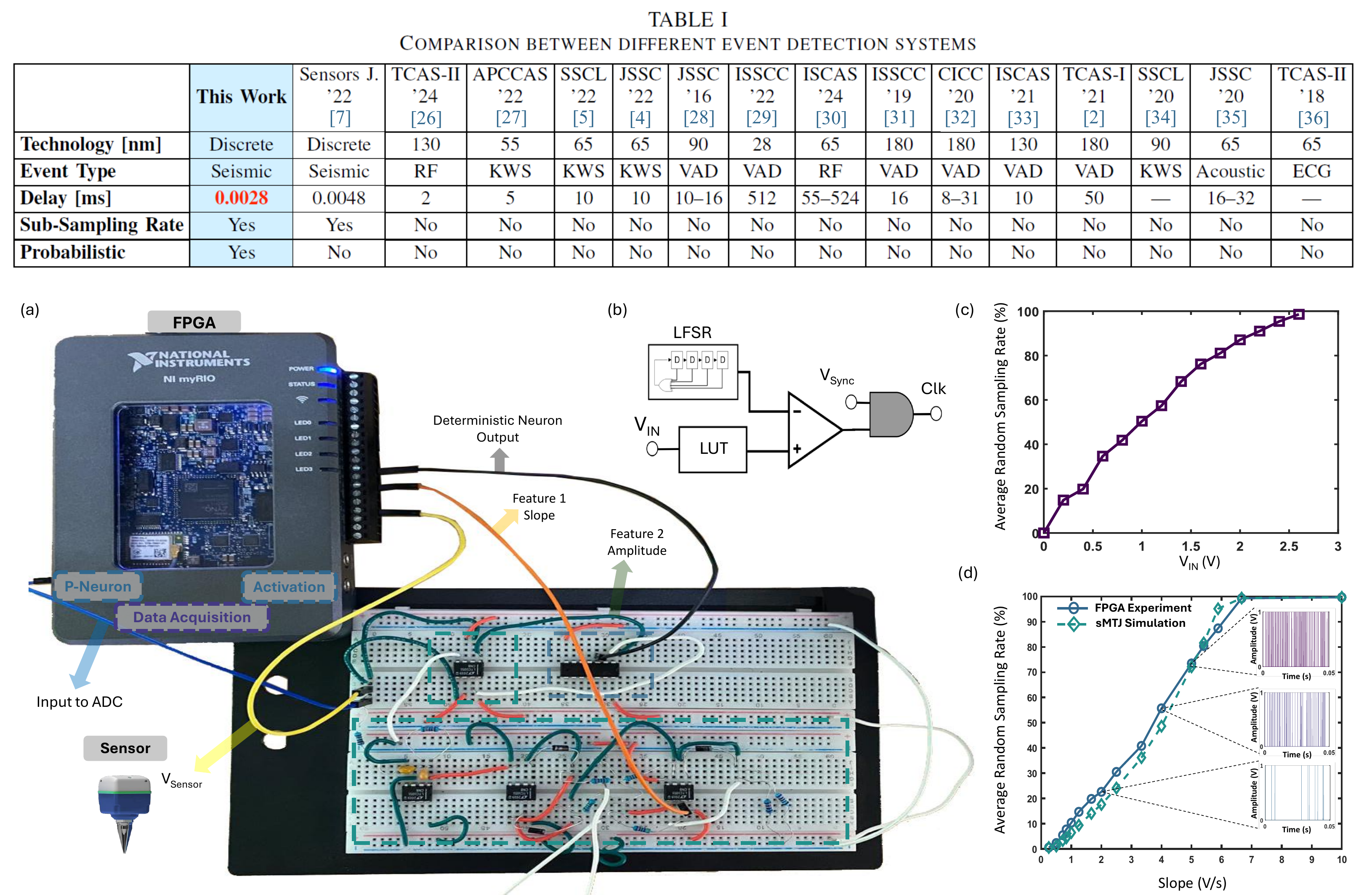}
    \centering
    \caption{\textbf{Experimental implementation of the probabilistic sensing system}, (a) using a field programmable gate array (FPGA) and discrete electronic components. (b) The design of the digital p-neuron implemented using the FPGA. Experimental results showing the average random sampling rate plotted against (c) the input signal to the p-neuron ($V_{IN}$) and (d) the slope of the original sensor signal. }
    \label{FPGA}
\end{figure*}

The overall architecture of the system is shown in Fig. \ref{MTJ} (a) - (c). The system is composed of three blocks; the Analog Feature Extraction (AFE) unit, the activation unit (including the p-neuron), and the data acquisition unit (including the Analog to Digital Converter (ADC)). The AFE unit \cite{our_paper,Sensor_Journal} extracts features from the analog sensor signal (slope and amplitude), and then uses such features to drive the input of the p-neuron in the activation unit, whose output is then used to control the activation of the ADC. For the slope, half-wave rectification followed by multiplication and subtraction is used. Two designs are presented, one based on a spintronic sMTJ p-neuron, and the other based on a digital FPGA-based p-neuron. The first is verified using simulations (as per \cite{Saleh_p-sensing}), while the latter is validated experimentally (Fig. \ref{FPGA} (a)). In the spintronic design, an isotropic magnet is used for the free layer magnet in the sMTJ (Fig. \ref{MTJ} (d)) due to its uniform $m_z(t)$ distribution \cite{Quantitative_Evaluation,adaptiveTRNGArXiv,PatentAdaptiveTRNG}, as can be seen in Fig. \ref{MTJ} (f), (g). Such a distribution is also possible using other technologies \cite{koh2025closed,AFMTRNG,SNWMemristor,RSQMemristorScaling}.

\section{Probabilistic Activation Using p-Neurons}

The p-neuron activates data acquisition at the onset of a definite event, and keeps all system blocks - including the data acquisition unit - in sleep-mode if no signature of an event is detected. However, in between these two extreme cases, if the event features are not prominent enough to declare a definite event, the p-bit would activate probabilistic random sampling instead. The degree of confidence in the detection of the onset of an event (i.e. the probability that a true event has been detected) is encoded into the time domain by controlling the average percentage of time the p-neuron activates data-acquisition, reflected in the average random sampling rate.

\section{Stochastic MTJ Retention Time}

The MTJ retention time ($\tau$) describes how long the sMTJ remains in a particular state on average, and hence determines how long the p-neuron state is retained if the input to it is unchanged.
The use of sMTJs with a long retention time - relative to the sampling interval of $V_{Sync}$ - makes the decision to sample or not to sample at any point in time always correlated to previous or upcoming sampling decisions (Fig. \ref{MTJ} (h)). 
On the other hand, p-neurons with faster sMTJs (with a retention time comparable to the sampling interval of $V_{Sync}$ (Fig.~\ref{MTJ}(i)) ensure that for every potential sampling instance the sampling decision is independent of other nearby sampling decisions.
Experimentally it has been demonstrated that $\tau$ can be varied from years down to nanoseconds \cite{Retention_time_2} through fine sMTJ engineering during fabrication, but this will require appropriate co-design of the p-neuron's OpAmp slew rate  \cite{Mohammed_2025}.

\section{Modular p-Neuron for Enhanced Control}
In order to control the average random sampling probability within the probabilistic range of operation the stochastic response of the p-neuron needs to be tuned \cite{Quantitative_Evaluation}. Furthermore, it would be useful to control the minimum average sampling rate during no-event periods - illustrated and denoted as ($X$) in Fig. \ref{MTJ} (j), (k). Controlling the average random sampling rate is fundamental for Probabilistic Sensing, while controlling the minimum average sampling rate benefits applications with interest in post-event information and background noise monitoring. Accordingly, it is desirable to have fine tunability over the probabilistic output of the p-neuron, and therefore, a modular p-bit design \cite{PatentPNeurons,PatentDecoupledPBit,NMDC} (modified version of \cite{Implementing_pbits}) with enhanced input-driven tunability, shown in Fig. \ref{MTJ} (b), has been employed for implementing the p-neuron in the presented system, shown in Fig. \ref{MTJ} (a) - (c). In the modified p-bit, the input voltage is applied directly to the comparator, while the gate voltage of the transistor is kept constant ($V_{REF}$). 

Finally, To ensure seamless integration with conventional systems, the output of the p-bit is ANDED with a synchronous clock signal ($V_{Sync}$) (Fig. \ref{MTJ}(b)) and the output is supplied to the clock of the ADC. Fig. \ref{FPGA} (c) validates the p-neuron’s ability in controlling the average sampling rate of the ADC through sweeping the input voltage ($V_{IN}$) of the p-neuron.

\begin{figure*}[!t]
\includegraphics[width= 6.6 in]{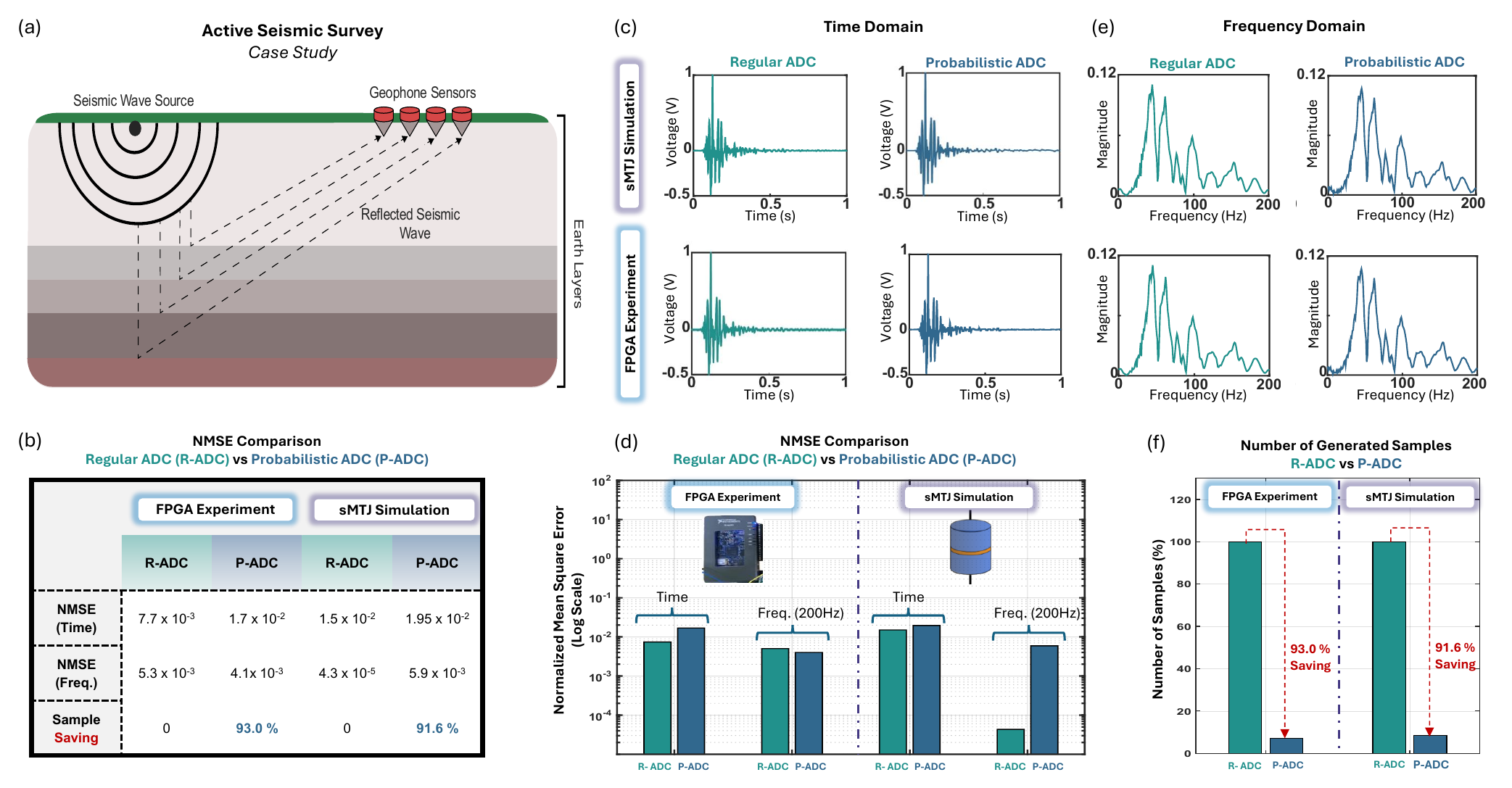}
    \centering
    \caption{Survey-level validation of the probabilistic sensing system using active seismic geophone sensor data. (a) schematic illustration on an active seismic survey. (b) Survey-level results based on both implementations; digital FPGA-based (experiment) and 
    spintronic sMTJ-based (simulation). Comparison of reconstructed seismic signals using linear interpolation after sampling using a regular ADC (green) and a probabilistic ADC (blue), (c) in the time domain and (e) in the frequency domain (frequency range 0 - 200 Hz). (d) Comparisson of the normalized mean square error (NMSE) between data acquired using an R-ADC and a P-ADC, for both the FPGA-based experiments and the sMTJ-based simulations. (f) Number of generated samples using the P-ADC in percentage relative to regular continuous sampling at a 2 kHz sampling frequency using an R-ADC.
    }
    \label{Results}
\end{figure*}

\section{Feature Extraction at Sub-Sampling Speed}

As can be seen in Fig. \ref{MTJ} (b), the input that controls the p-neuron is supplied from the AFE unit. 
Information in signals is embedded in a variety of features, and the optimal feature - or combination of features - for event detection vary depending on the application of interest. In our system we extract two event features: (1) amplitude, and use it to trigger deterministic activation, and (2) rate of change, and use it to trigger probabilistic activation. We implement our system and test it on data from active seismic surveys \cite{survey_4, real_data}, where the signal's rate of change can be considered as a prominent signature of the onset of an active seismic event (also applicable to many acoustic signals). Fig. \ref{FPGA} (d) shows how the rate of change of the signal (its slope) is used to control the average random sampling rate. Table I shows a comparison between the system's response time and the response time of other event detection systems reported in the literature \cite{Feature_4, AFE, AFE2, Sensor_Journal, newref1, ref16, ref20, ref21, newref2, ref23, ref24, ref25, ref27, ref28, ref29},
showing that the response time of event-detection in the presented systems is on the order of microseconds (2.8 $\mu$ sec), while other systems are on the order of milliseconds, allowing them to achieve post-event detection only, and not real-time event detection that can be used to activate data acquisition. 

The limit on the required speed of event detection is dictated by the sampling frequency used in data-acquisition, where event-detection speed needs to be faster than the sampling period between samples, which we refer to here as the sub-sampling rate limit. Our system is able to break the sub-sampling rate speed limit because the detection mechanism is specialized using only two event features that are hard wired into the system (like reflex responses in the nervous system), and takes place entirely in the analog domain.

\section{Probabilistic Active Seismic Sensing}

The system was tested using geophone sensor data from an active seismic survey \cite{real_data}. A sample seismic event signal is shown in Figs. \ref{Results} (c). In active seismic sensing the typical sampling rate is around 2 kHz \cite{survey_4}, and hence the sMTJ’s retention time ($\tau$) was set to 500 $\mu$s. The results in Fig. \ref{MTJ} (j), (k) show the output of the activation unit in response to an active seismic signal, and demonstrate the system's ability in achieving real-time probabilistic event detection. The results also demonstrate how $V_{REF}$ can be used to tune the average minimum random sampling rate (X) during no-event periods.

To characterize the accuracy of the acquired data using the proposed Probabilistic Sensing approach, the entire probabilistic data-acquisition system (Probabilistic ADC) was tested using the data of the first 50 active seismic events from the survey and compared with a regular continuous sampling ADC (Regular ADC) tested using the same data, in both simulation and FPGA experiments. The data obtained using both systems, the regular ADC (R-ADC) and the probabilistic ADC (P-ADC), was compared in both time and frequency domains. The results verify that both time and frequency domain plots are almost identical, as shown in Figs. \ref{Results}(c) and (e), respectively, confirming the feasibility of lossless Probabilistic Sensing, where the same information was acquired using the P-ADC when compared with the continuous-sampling R-ADC.

In order to further quantitatively compare both schemes, the reconstructed signals were bench-marked with the original sensor signals for all the active seismic events in the survey. The two sampling schemes were quantitatively evaluated based on the normalized mean squared error (NMSE) between the original raw seismic signal and the reconstructed data - using linear interpolation - after sampling.
Table II (Fig. \ref{Results} (b)) compares the NMSE of all signals reconstructed using both ADCs in time and frequency domains. For all evaluated NMSEs, the differences between both schemes are minimal (Fig. \ref{Results} (d)). The experimental NMSE in the 0 - 200 Hz frequency range (application critical range) indicates that the error using the proposed sensing approach can be as small as 0.41 $\%$, comparable to regular continuous sampling using the R-ADC. Most importantly, the results (Fig. \ref{Results} (f)) further demonstrate that probabilistic sensing can achieve loss-less data acquisition with great savings in the number of generated samples when compared with the R-ADC. In all events investigated, a 93 $\%$ saving in the number of generated samples and in the active operation time of the ADC were achieved.

\section{Conclusion}
We presented the concept of probabilistic sensing via p-neurons and demonstrated its feasibility by designing and implementing a probabilistic sensing system that achieves probabilistic event-based data sampling; enabling intelligence in data acquisition. The system was tested on data from an active seismic survey demonstrating loss-less probabilistic data-acquisition with an NMSE of 0.41 \% and a 93 \% saving in the number of generated samples.

\IEEEtriggeratref{19}
\bibliographystyle{IEEEtran}
\bibliography{library}

\end{document}